# Language, communication and society: a gender based linguistics analysis

P. Cutugno, D. Chiarella, R. Lucentini, L. Marconi and G. Morgavi

***Abstract*—**The purpose of this study is to find evidence for supporting the hypothesis that language is the mirror of our thinking, our prejudices and cultural stereotypes. In this analysis, a questionnaire was administered to 537 people. The answers have been analysed to see if gender stereotypes were present such as the attribution of psychological and behavioural characteristics.

In particular, the aim was to identify, if any, what are the stereotyped images, which emerge in defining the roles of men and women in modern society. Moreover, the results given can be a good starting point to understand if gender stereotypes, and the expectations they produce, can result in penalization or inequality. If so, the language and its use would create inherently a gender bias, which influences evaluations both in work settings both in everyday life.

## I. Introduction

LANGUAGE transmits information, in number and variety, far more than appears on the surface. There is a direct relationship between reality, language and thought. In particular, the language expresses our vision of reality: it does not reflect the world in itself, but the way it is interpreted by us. The language reflects the culture of a society and influences its behaviours. Supporting the gender language, in order to sensitize society on the proper use of the Italian language in a respectful perspective of both genders, is an important step to address the problem of violence against women. Violence against women is an atrocious violation of fundamental rights. In Italy, in 2011, one hundred thirty-seven women were murdered; in 2012 one hundred twenty-four and, in 2013, there were one hundred thirty-four victims of femicide. In Italy, sex crimes have been thirty-six in the first six months of 2014. In the majority of cases, women were killed by their husbands, partners or former partners.

The first thematic report on femicide was made in Italy by Rashida Manjoo and was presented to ONU in 2012 [1]. In this report it was pointed out that most of the violence is not reported because domestic violence is not recognized as a crime and victims are economically subordinate than the perpetrators of violence. In particular, the report pointed out that, in Italy, gender stereotypes are deeply rooted and predetermine the roles of men and women in society.

To deal with these issues, the report “on eliminating gender stereotypes in the EU" [2] highlights the need to:

- Emphasise the need for education programmes/curricula focusing on equality between men and women, respect for others, respect amongst young people, respectful sexuality and rejection of all forms of violence, as well as the importance of training teachers in this subject;
- Emphasise the need for a gender mainstreaming process in schools and therefore encourages schools to design and implement awareness training exercises and practical exercises designed to promote gender equality in the academic curriculum;
- Point out that, although a majority of countries in the EU have gender-equality policies for higher education, almost all the policies and projects are focused on young women; calls, therefore, on the Member States to draw up general national strategies and initiatives combating gender stereotyping in higher education and targeting young men.

Although any specific course focused on respect for gender and gender equality is not envisaged by the Italian government, in Italy we find countless projects implemented by regions [3], provinces [4], schools [5] and community services sector [6]. All these projects are aimed at raising awareness of these problems among teenagers and the public. This preliminary work aims the same objectives by providing a basic survey on gender stereotypes [7] [8] in Italian language.

## II. The analysis

### A. Data set

In order to reflect on "being men and women" and "building positive relationships", the City of Genoa (Italy) has distributed a brief questionnaire (see Figure 1 Questionnaire submitted to the citizens), in the period from November 2013 to February 2014, where it was asked to reflect about stereotypes and prejudices and how they are accomplices of violence.

D. Chiarella is with the National Research Council of Italy, Institute of Intelligent Systems for Automation – Genoa branch, Genova, Italy (phone: +39-010-6475220; fax: +39-010-6475207; e-mail: chiarella@ge.issia.cnr.it or davide.chiarella@ilc.cnr.it).

P. Cutugno, R. Lucentini, L. Marconi are with the National Research Council of Italy, Institute of Computational Linguistics “A. Zampolli” – Genoa branch, Genova, Italy (e-mail: paola.cutugno | roberta.lucentini | lucia.marconi @ilc.cnr.it).

G. Morgavi is with the National Research Council of Italy, Institute of Electronics, Computer and Telecommunication Engineering – Genoa branch, Genova, Italy. (e-mail: giovanna.morgavi@ieiit.cnr.it)

In Fig. 1, the structure of the questionnaire given to the citizens of the City of Genoa is shown. The questionnaire was made up of the following questions:

1. three words about what you think is important in a loving relationship;

2. three adjectives to define "masculine";

3. three adjectives to define "female";

4. three things that hurt in a loving relationship;

5. free thoughts.

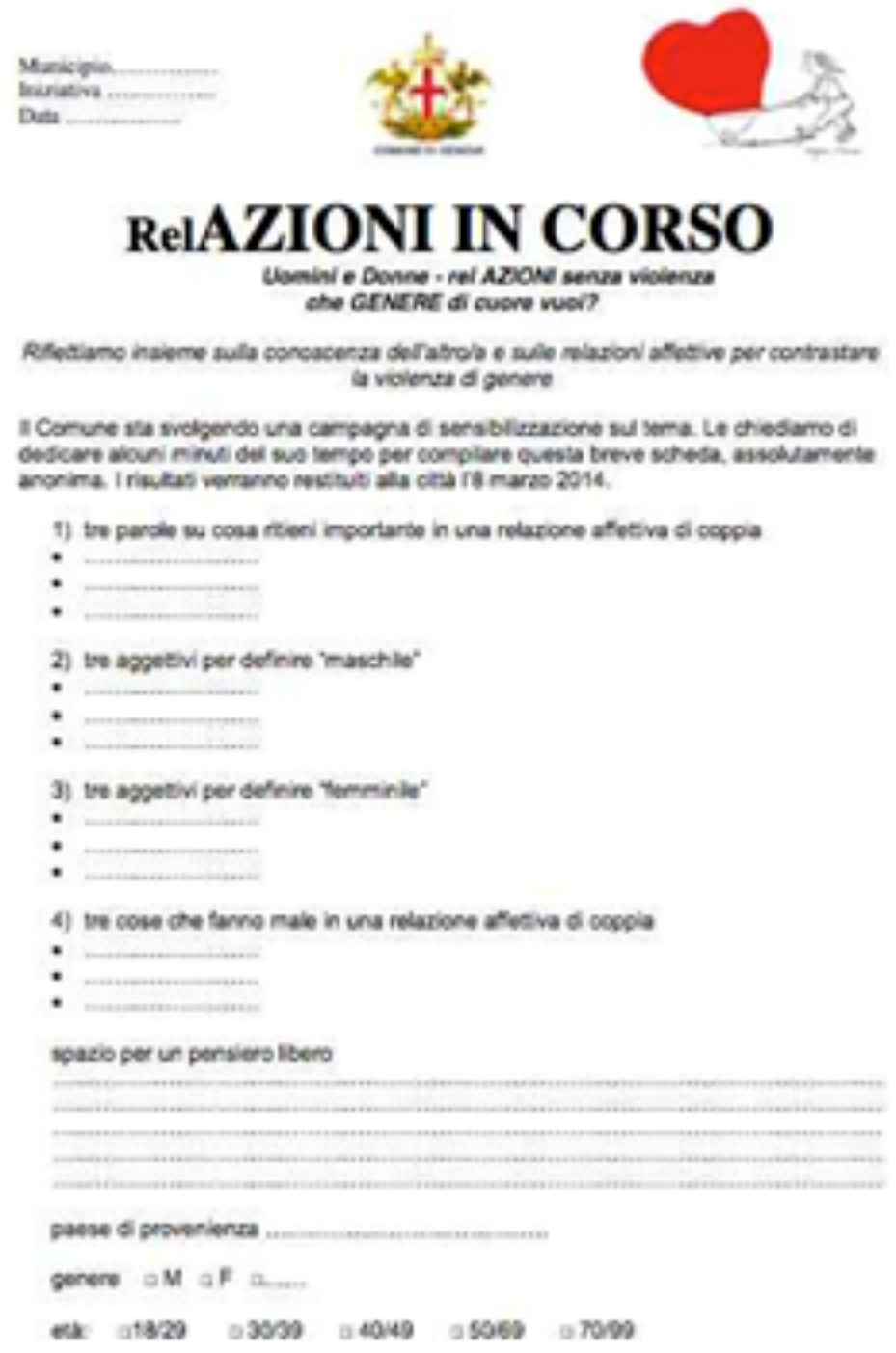
Municipio..............
Iniziativa ..............
Data ..............

RelAZIONI IN CORSO

*Uomini e Donne - rel AZIONI senza violenza*
*che GENERE di cuore vuoi?*

*Riflettiamo insieme sulla conoscenza dell'altro/a e sulle relazioni affettive per contrastare la violenza di genere*

Il Comune sta svolgendo una campagna di sensibilizzazione sul tema. Le chiediamo di dedicare alcuni minuti del suo tempo per compilare questa breve scheda, assolutamente anonima. I risultati verranno restituiti alla città l'8 marzo 2014.

1) tre parole su cosa ritieni importante in una relazione affettiva di coppia
- ..............
- ..............
- ..............

2) tre aggettivi per definire "maschile"
- ..............
- ..............
- ..............

3) tre aggettivi per definire "femminile"
- ..............
- ..............
- ..............

4) tre cose che fanno male in una relazione affettiva di coppia
- ..............
- ..............
- ..............

spazio per un pensiero libero
..............

paese di provenienza ..............

genere □M □F □......

età: □18/29 □30/39 □40/49 □50/69 □70/99

**Figure 1 Questionnaire submitted to the citizens**

The short questionnaire has been proposed to the following age groups: 18/29 - 30/39 - 40/49 - 50/59 - 70/99. It was compiled in manuscript form by 334 people and in web form by 203 people. The questionnaire has been completed by 459 women, 75 men and 3 transgender for a total of five-hundred thirty-seven people.

**Table I Groups of respondents divided by age**

| 18/29 | 30/39 | 40/49 | 50/59 | 70/99 |
|---|---|---|---|---|
| **220** | 66 | 68 | 155 | 17 |
| ***11 people cannot be classified because they didn't provide their age** | | | | |

There is a big difference in numbers between women who completed the questionnaire, compared to men.

This big difference, of course, leads us to reflect and to ask: what does this mean?

Perhaps it is true that women are more numerous in Italy, but it is necessary to point out that women are more sensitive and willing to deal with these problems than men are.

This work, based on the answers given to the questionnaire, tries to understand if stereotypes were present in the answers of the interviewees. In particular, the main aim is to identify what are the "stereotypical" images to define the roles of male and female. The answers have been analysed from a linguistic point of view, in order to verify if the language reveals the perception of reality through images, concepts and beliefs that exist in our minds, and if the language is revealing of stereotypes.

### *B. What is a stereotype?*

A stereotype is a set of beliefs, simplistic representations, and oversimplified views of reality rigidly connected to each other, that a social group associated with another group. Stereotypes are born by the behaviour of a group of people or a gender (masculine or feminine). Therefore, gender stereotypes are a subclass of stereotypes.

The examples seem trivial, but it is not so, because stereotypes not only affect the ideas of groups of individuals, but also have implications in our actions and in our society.

For example, there are stereotypes associated with the rigid division of roles in the family and in the social and professional sphere.

A key feature of the language is to convey information in quantity and variety much bigger than we imagine. It is, indeed, through language that our vision of reality and society is transmitted to others; the spoken or written language is therefore an important vehicle of common sense [9].

### *C. First question*

The first question was "three words about what you think is important in a loving relationship". The most frequent answered words for group 18-29 (the largest group) have been:

- "Fiducia" ["trust"]: 127 occurrences.
- "Rispetto" ["respect"]: 97 occurrences.
- "Amore" ["love"]: 80 occurrences.
- "Sincerità" ["sincerity"]: 68 occurrences.

Given these most frequent words, triplets referring to the same age group (i.e. 18-29) containing at least one of them were extracted in order to see if and how people in the same age group had used the same set or subset of words.

The same procedure was conducted for all age groups in order to identify the triplets of most frequent words. Then, the most frequent words of each age group were compared to see if:

- There was a same set of words used in different age groups.
- Words were the same or were similar.

Fig. 2 shows the set of words that contain at least one of the four most frequent words of Table II related to the age group 18 -29; e.g., "fiducia", "amore", "rispetto", "sincerità" ["trust", "love", "respect", "sincerity"].

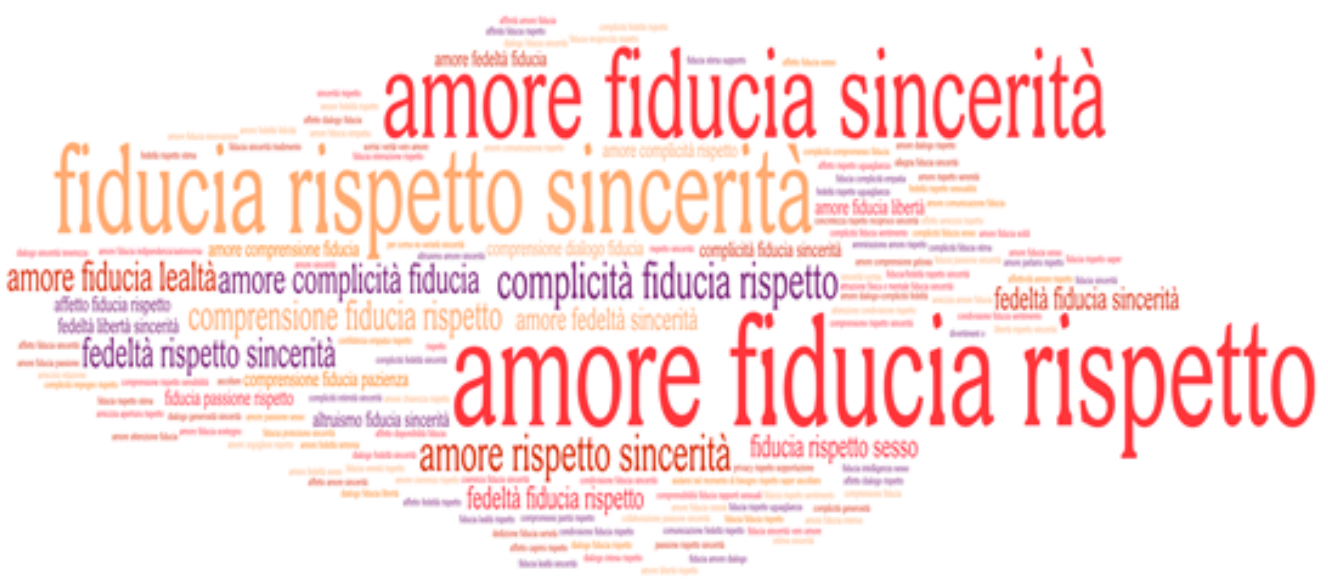

**Figure 2: triplets for first question, age group 18-29**

The greater or lesser font size of characters indicates the frequency degree of use of the triplet. Therefore, Fig. 2 shows the more or less common use of the same words used in response to the first question by people of the same age.
Note that the most common triplet is “amore, fiducia, rispetto” ["love, trust, respect"] and then, in descending order of frequency, “fiducia, rispetto, sincerità” ["trust, respect, sincerity"] and “amore, fiducia, sincerità” ["love, trust, sincerity"] and so on.
Furthermore, the most frequent word is “fiducia” ["trust"] that is linked with the words: “amore e rispetto” ["love and respect"], “rispetto e sincerità” ["respect and sincerity"], “amore e onestà” ["love and honesty"], “complicità e rispetto” ["complicity and respect"], “comprensione e rispetto” ["comprehension and respect"], “amore e complicità” ["love and complicity"], “amore e fedeltà” ["love and faithfulness"], etc.

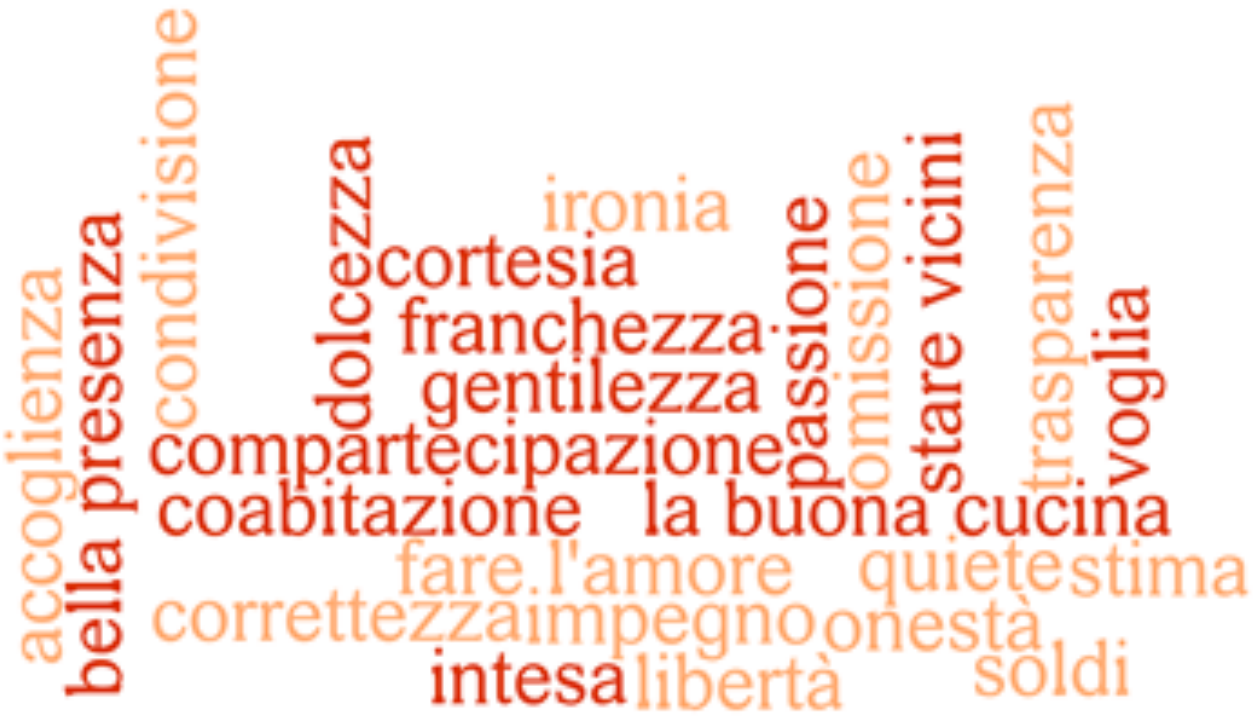

**Figure 3: top rated single words for first question for age group 18-29 (lighter) and 50-69 (darker), male respondents**

Fig. 3 and 4, which refer to the first question of the questionnaire, show the words used in the two age groups 18-29 and 50-69 from the only male gender (M) and female gender and male (FM). In both figures, the words written with a darker font refer to the age group 50-69.
In Fig. 3, the words quoted by the male gender in the age group 18-29 are: “accoglienza, condivisione, correttezza, fare l’amore, impegno, ironia, libertà, omissione, onestà, quiete, soldi, stima, trasparenza” ["acceptance, sharing, fairness, making love, commitment, humor, freedom, omission, honesty, peace, money, respect, transparency"].

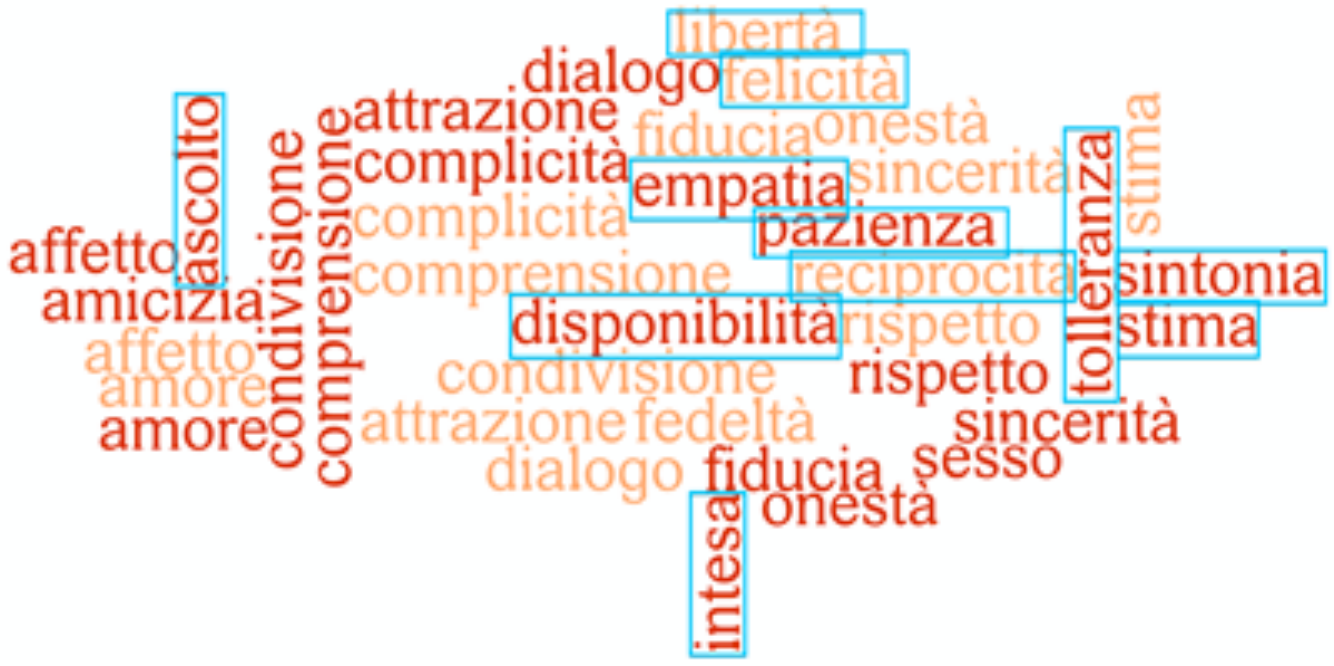

**Figure 4: single words in common for first question between age group 18-29 (lighter) and 50-69 (darker), male and female respondents**

Instead, the words quoted by the male gender for the age group 50-59 are: “bella presenza, coabitazione, compartecipazione, cortesia, dolcezza, franchezza, gentilezza, intesa, la buona cucina, passione, stare vicini, voglia” ["beautiful presence, cohabitation, sharing, kindness, gentleness, honesty, kindness, understanding, good food, passion, stay close, desire"]. As it can be seen, in different age groups, the male gender has used different words that can show the evolution of thinking through aging in men.
Fig. 4 contains the top rated words related to both genders (FM) for the same age groups. The words used are the following for the age group 18-29: “affetto, amore, attrazione, complicità, comprensione, condivisione, dialogo, fedeltà, felicità, fiducia, onestà, reciprocità, rispetto, sincerità, stima” ["affection, love, attraction, complicity, comprehension, sharing, dialogue, faithfulness, happiness, trust, honesty, reciprocity, respect, sincerity, esteem"]. The words of the age group 50-59 are: “affetto, amicizia, amore, ascolto, attrazione, complicità, comprensione, condivisione, dialogo, disponibilità, empatia, fiducia, intesa, onestà, rispetto, sesso, sincerità, sintonia, stima, tolleranza” ["Affection, friendship, love, listening, attraction, complicity, comprehension, sharing, dialogue, willingness, empathy, trust, understanding, honesty, respect, sex, sincerity, harmony, esteem, tolerance"].
In this case, there are some words in common to the two age groups. The words “felicità, libertà, reciprocità” [“happiness, freedom, reciprocity”] are present only in the age group 18-29 and the words “ascolto, disponibilità, empatia, intesa, pazienza, sintonia, stima, tolleranza” ["listening, willingness, empathy, understanding, patience, harmony, respect, tolerance"] are present only in the age group 50-69. The diversity of words is explanatory of what is considered important in a loving relationship from two different generations both females both males.

**Table II - First question: highest frequency words per age groups**

| 18-29 | 30-39 | 40-49 | 50-69 | 70-99 |
|---|---|---|---|---|
| **Trust** | Respect | Respect | Respect | Respect |
| **Respect** | Faithfulness | Love | Love | Sincerity |
| **Love** | Love | Trust | Trust | Comprehension |
| **Sincerity** | Complicity | Sharing | Complicity | Love |
| **Faithfulness** | Sincerity | Complicity | Sharing | Feelings |

If we analyse the most answered words to the first question, it can be observed that there is a certain homogeneity between the age groups (see Table II). The most frequent words, in each age group, are the words “rispetto” and “amore” ["respect" and "love"]. The word “fiducia” ["trust"] appears in almost all the age groups, except in the group of 70-99 years old that, however, being a very small sample, it cannot be considered representative for the study.

### *D. Second question*

In the second question, the word on which the attention is focused is an adjective (male); the adjectives are used to give meaning and to classify the words; in general, they are a sample of our thinking [9] [10].

Adjectives used by interviewees were 1475; the ones given by those people who have not declared either gender or age were excluded. The largest sample was of 601 adjectives related to the age group 18 - 29. Some of the answers, unfortunately, do not belong to the grammatical category requested (i.e. the adjective), anyway it has been chosen to analyse all the answers even if they were not adjectives.

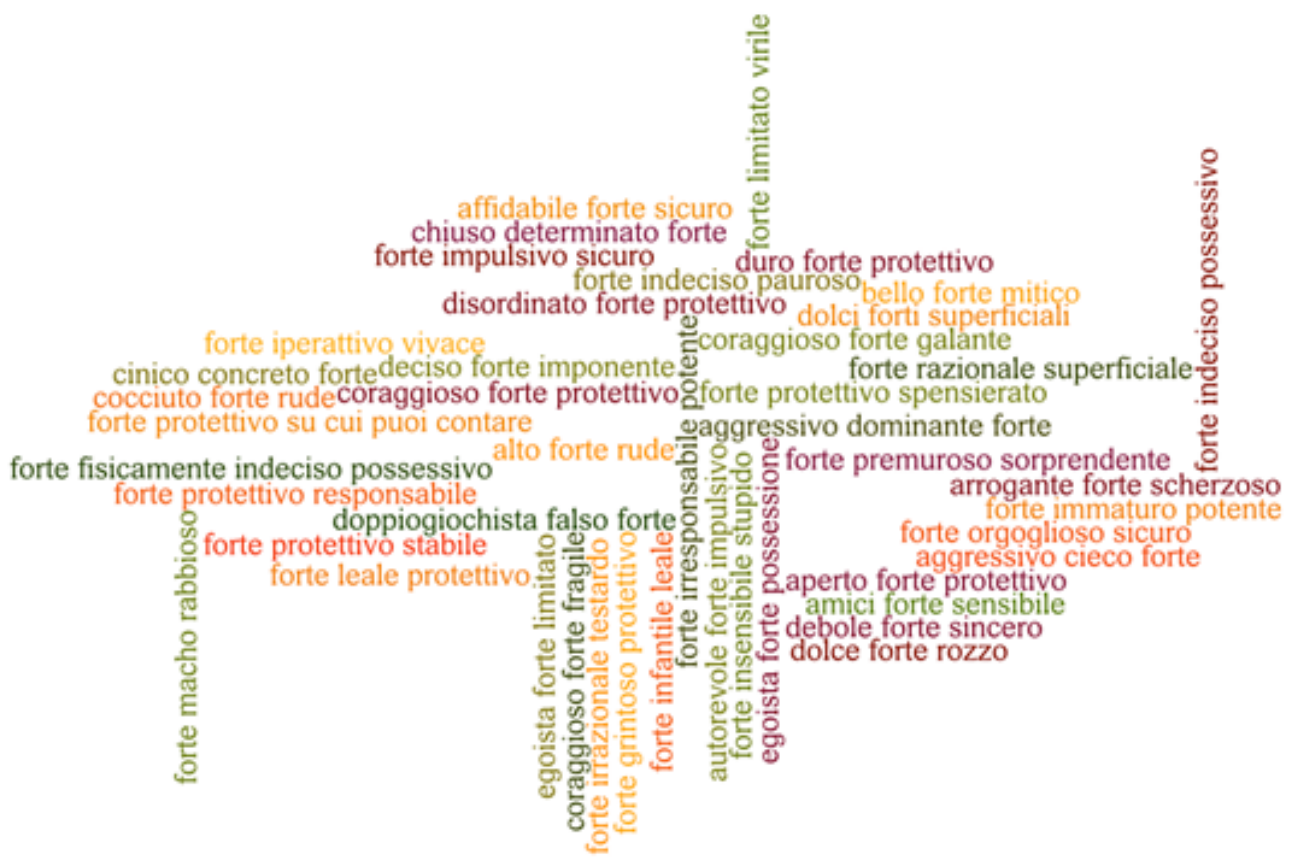

**Figure 5: triplets for second question, age group 18-29**

The adjective with the highest number of occurrences is "forte" [“strong”]. The second word was "protettivo" [“protective”] and the third "possessivo" [“possessive”] at the same frequency with "egoista" [“selfish”]. The analysis of the age group 30-39 has revealed 187 adjectives; the largest number of occurrences was reserved for the word “protettivo” ["protective"] followed by “forte” ["strong"]. The age group 40-49 used 22 adjectives. In this age group, the female gender expressed a preference for the adjective “forte” ["strong"] followed by “protettivo” ["protective"] and “egoista” ["selfish"]. The age group 50-69 expressed a preference for the word “forte” ["strong"], followed by “intelligente” ["intelligent"] and “protettivo” ["protective"]. The lexeme with the largest number of occurrences attributed to the concept of masculinity is the adjective “forte” ["strong"]. It is used especially for the female gender in almost all age classes, showing how the stereotype of the "stronger sex" is rooted in our society.

The use of gender stereotypes lead to a rigid and distorted perception of reality, which is based on what we mean by "feminine" and "masculine" and what we expect from women and men. Through this way of thinking, a priori expectations on the roles, which men and women should take in society, are established only because of being biologically male or female. For example, a woman is considered quieter, less aggressive, good listener; a woman loves to take care of others, while man has a strong personality, great logical skills, spirit of adventure and leadership.

Fig. 6 shows the words used in the two age groups 18-29 and 50-69 by the female gender and male. The words written with the darker character refer to the age range 50-69.

The words used by female and male gender in the age group 18-29 are: “aggressivo, arrogante, coraggioso, concreto, dolce, egoista, forte, forza, fragile, galante, geloso, indeciso, infedele, intelligente, introverso, istintivo, megalomane, menefreghista, passionale, possessivo, protettivo, protezione, rozzo, sicuro, stabile, superficiale, virile” ["aggressive, arrogant, brave, pragmatic, sweet, selfish, strong, strength, fragile, gallant, jealous, hesitant, unfaithful, smart, introverted, instinctive, megalomaniac, uncaring, passionate, possessive, protective, protection, rude, self-confident, superficial, manly"].

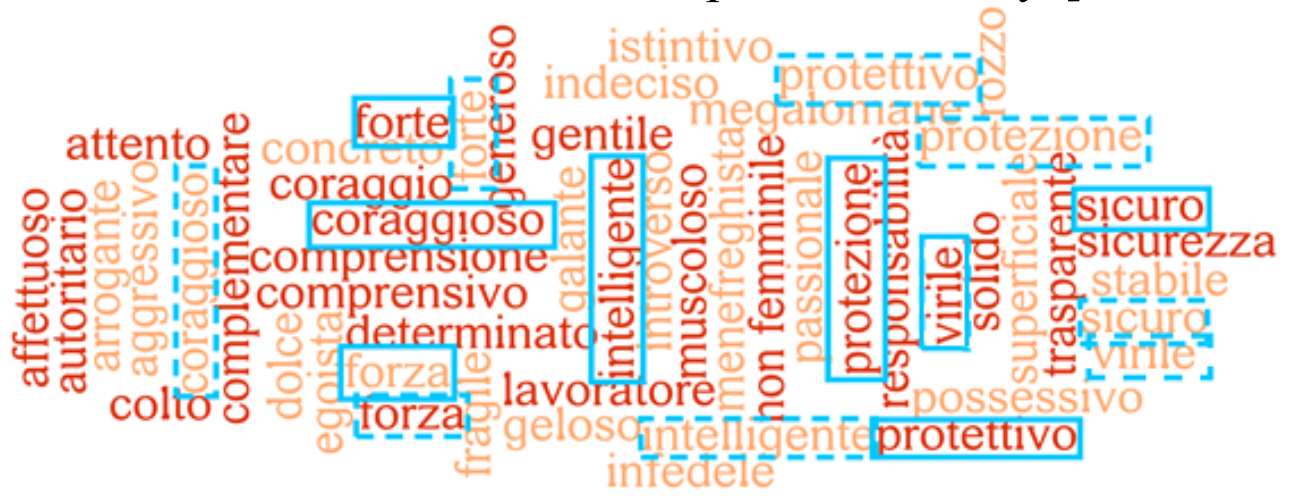

**Figure 6: Words for age groups 18-29/50-69 (both genders) which define a male**

The words used by female and male gender in the age group 50-69 are: “affettuoso, autoritario, attento, colto, complementare, comprensione, comprensivo, coraggio, coraggioso, determinato, forza, forte, generoso, gentile, intelligente, lavoratore, muscoloso, non femminile, protettivo, sicurezza, sicuro, solido, trasparente, virile” ["loving, authoritarian, careful, cultured, complementary, comprehension, understanding, bravery, brave, determined, strenght, strong, generous, kind, smart, hardworking, muscular, not feminine, protective, safety, self-confident, tough, transparent, manly"].

Some words are common to the two age groups and genders: in particular "courageous, strenght, strong, smart, protective, protection, self-confident, manly". These words are circled in blue in the Fig. 6 (dashed line for group 18-29, continuous for 50-69).

### *E. Third question*

As for the third question (i.e. three adjectives to define "female"), we can observe that the lexemes more used were those shown in Table III represented in the different age groups.

Since the female gender is the most represented in the questionnaire, it is our interest to focus our attention on the self-stereotype of women.

**Table III - Third question: highest frequency words per age groups**

| 18-29 | 30-39 | 40-49 | 50-69 |
|---|---|---|---|
| **Problematic** | Comfortable | Comfortable | Strong |
| **Comprehension** | Sensitive | Strong | Comfortable |
| **Efficient** | Sweet, Maternal | Sensitive | Sweet/sweetness Maternal |
| **Jealous** | Patient | Maternal | Sensitive |
| **Comprehensive Elegant** | Creative Unselfish | Dedicated/loving Sweet Empathic | Willing |

Fig. 7 shows, for the most representative groups, how the self-stereotype of women is to be “comprensiva, di occuparsi degli altri, sensibile, dolce e materna” [“comprehensive, to take care of others, sensitive, gentle and maternal”].

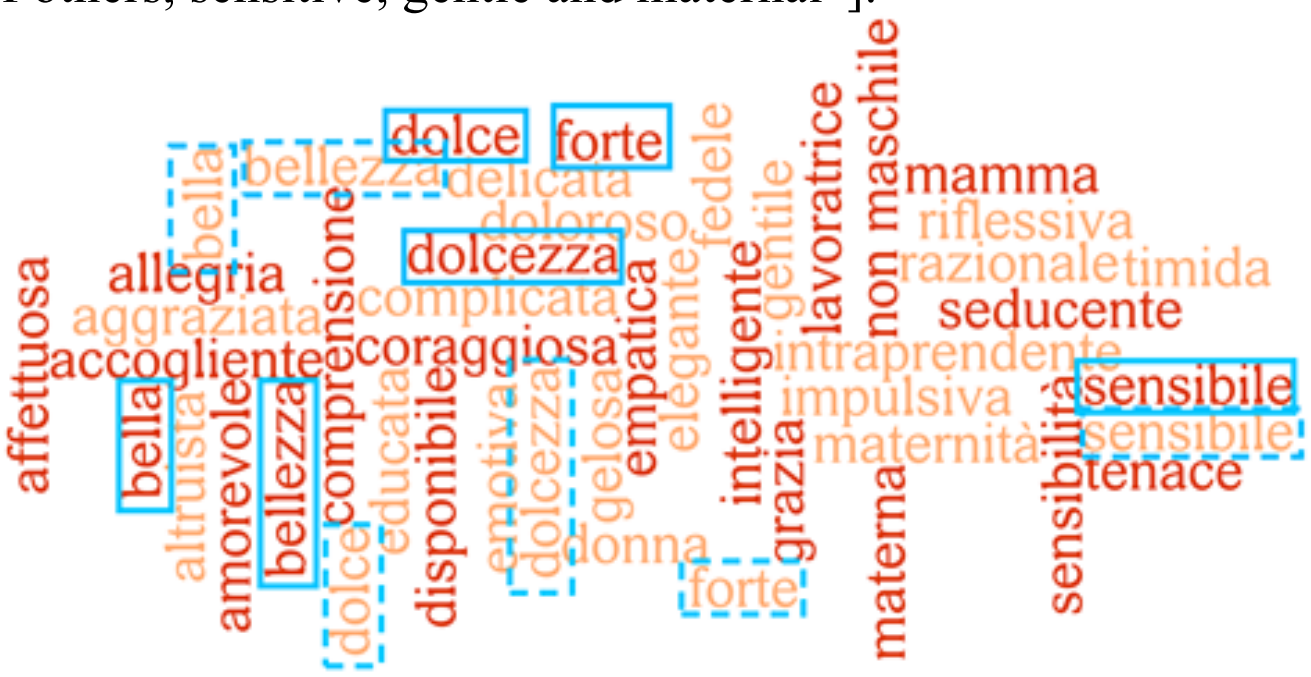

**Figure 7: Words for age groups 18-29 50-69 (both genders) which define a female**

Considering the set of the two questions related to the concepts of "male" and "female", a benevolent sexism in its peculiarity is revealed. In fact, in the first question, men are in charge of protecting women and are responsible for their protection and, in the second one, the image of the female in her private and emotional dimension is given: a social context, where women must provide and ensure the welfare of man.

## *F. Fourth question*

The fourth question asked to identify three things that hurt in a loving relationship. As for the previous questions the most representative groups were examined.

In the age group 18-29, the answers related to the fourth question consist of 197 sets of words used by the female gender and 19 sets of words used by the male gender.

As shown in Fig. 8, the sets of three words most frequently used to characterize the three things that hurt in a loving relationship are “falsità, tradimento, violenza” ["deceitfulness, unfaithfulness, violence"] followed by “bugie, tradimento, violenza” ["lies, unfaithfulness, violence"], etc.

Figure 9 shows the words used by age group 50-69. It's interesting to note that there were no particular variations on what is considered hurting by different age groups in a loving relationship.

In Fig. 9, indeed, we find the same sets of words: “falsità, tradimento, violenza” ["deceitfulness, unfaithfulness, violence"] followed by “bugie, tradimento, violenza” ["lies, unfaithfulness, violence"], etc.

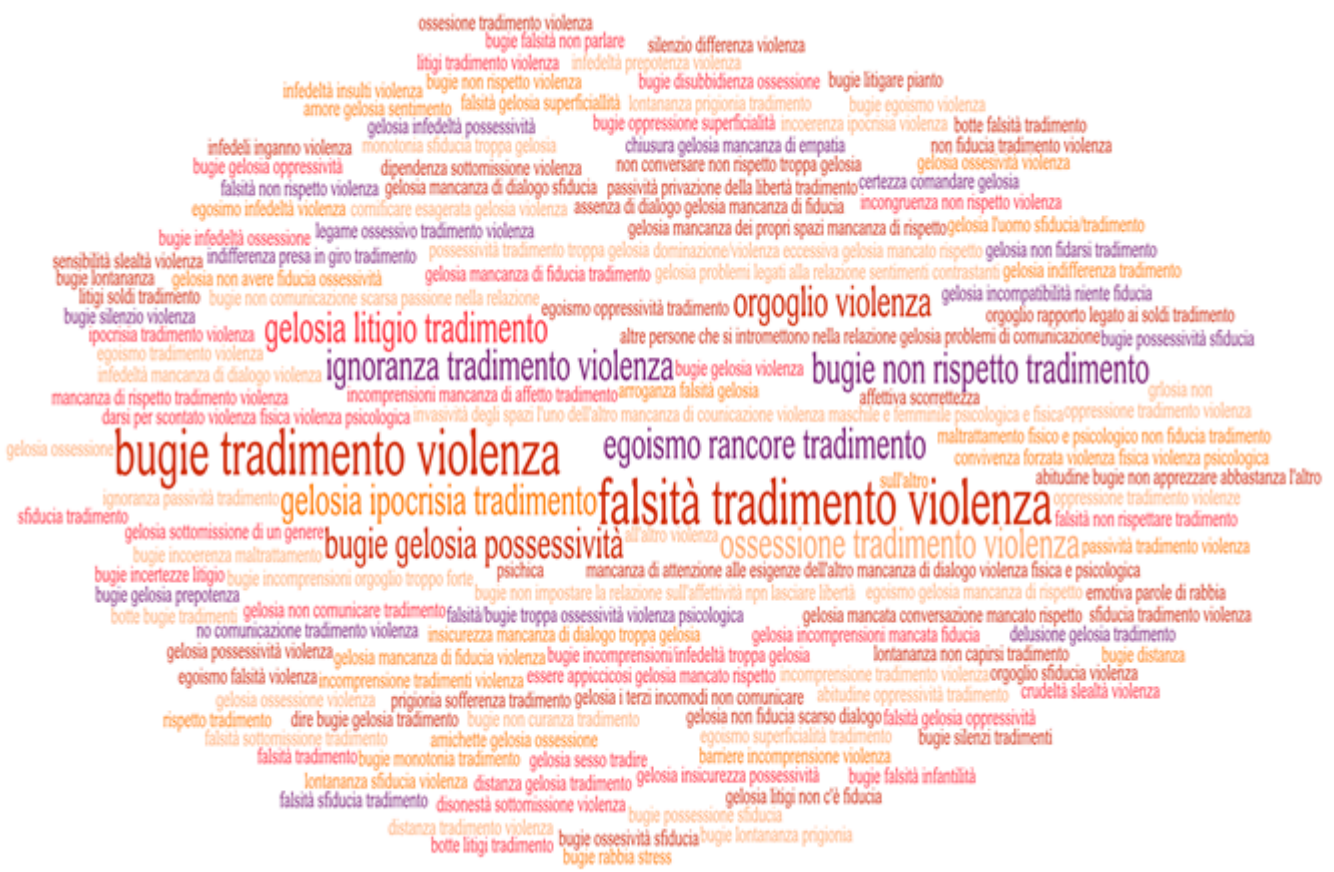

**Figure 8: triplets for fourth question, age group 18 -29**

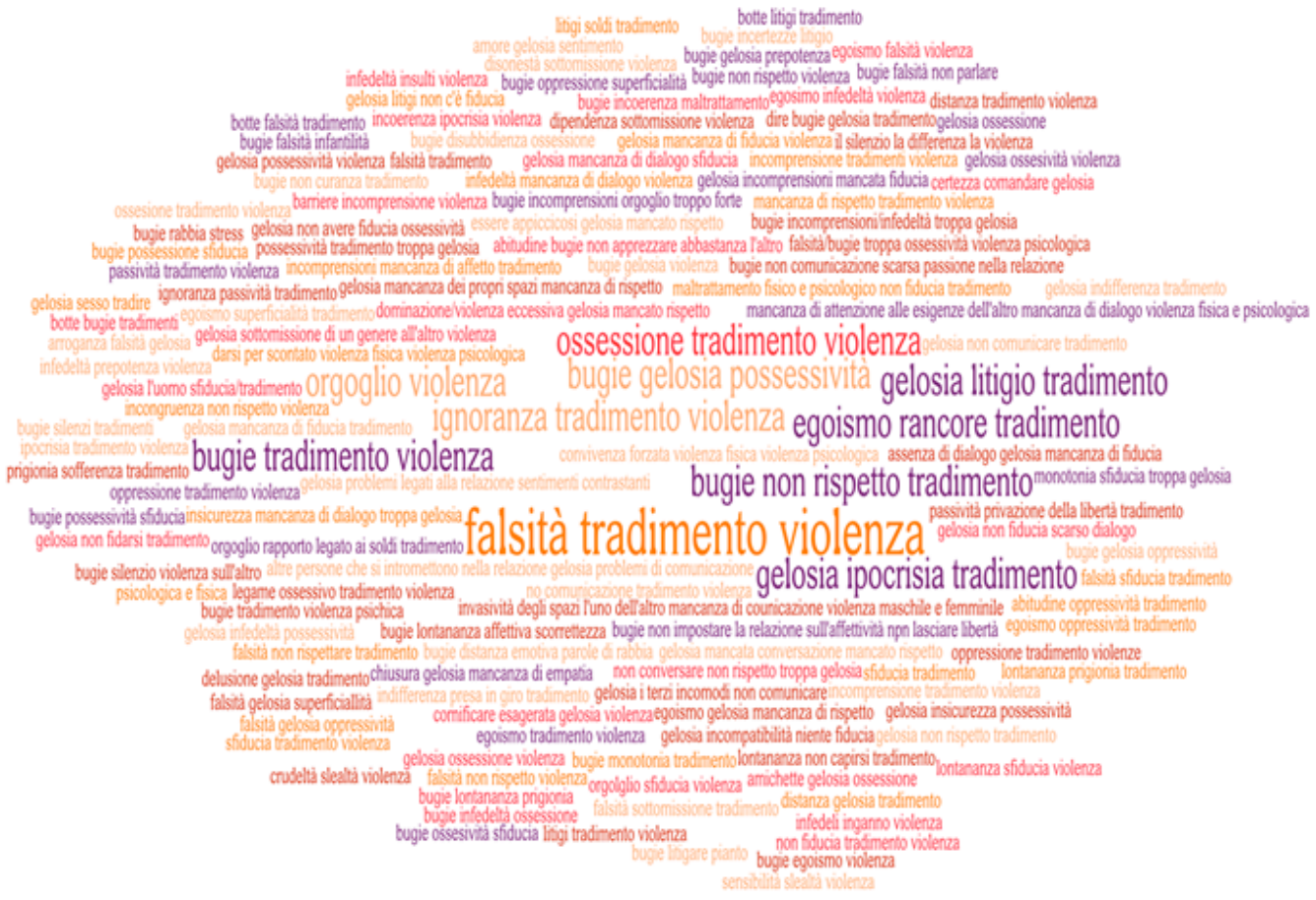

**Figure 9: triplets for fourth question, age group 50 -69**

In particular, in Table IV, it can be seen that the first word of each line is the one with the highest frequency within each age group. Then we find the words in descending order according to their number of occurrences; the last line of Table IV, characterized by a line more evident, contains the words close ,as number of occurrences, to the last ones for each age class.

**Table IV - Fourth question: highest frequency words per age groups**

| 18-29 | 30-39 | 40-49 | 50-69 | 70-99 |
|---|---|---|---|---|
| **Unf.*** | Jealousy | Selfishness | Violence | Jealousy |
| **Jealousy** | Selfishness | Apathy | Apathy | Lack of respect |
| **Violence** | Incomprehension | Silence | Jealousy | Unf.* |
| **Lies** | Violence, distrust, lack of respect | Violence Jealousy | Selfishness | Apathy |
| | Unf.* | Lies, unf.* | Lies, unf.* | Violence, lies |

Where Unf.* stays for Unfaithfulness

The words “gelosia” ["jealousy"] and “violenza” ["violence"] are present in all age groups. “Infedeltà” ["infidelity"] is synonymous with “tradimento” ["unfaithfulness"], for this reason we can add to the words “gelosia” ["jealousy"] and “violenza” ["violence"] the word "unfaithfulness" that is constant in the different age groups.

Therefore, data seem to indicate that “gelosia” ["jealousy"], “violenza” ["violence"] and “tradimento” ["unfaithfulness"] are among the things considered the most harmful in a loving relationship.

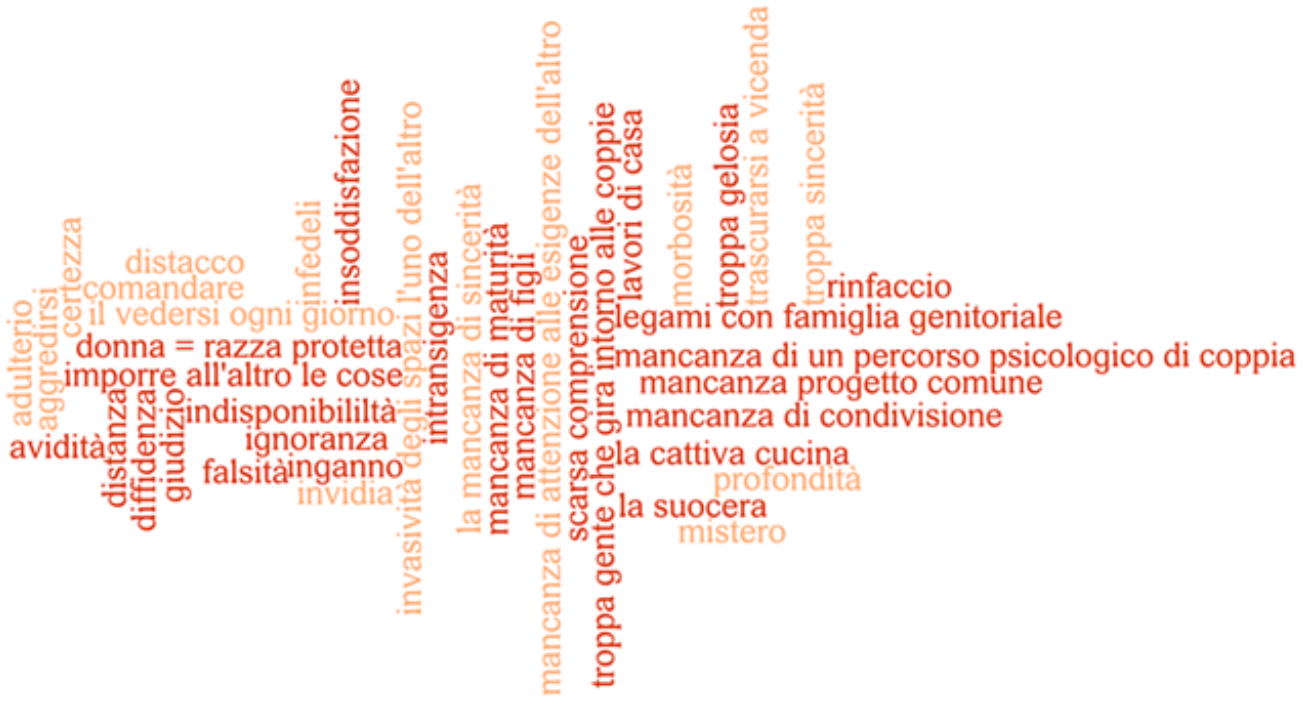

**Figure 10: top rated single words for fourth question for age group 18-29 (lighter) and 50-69 (darker), male respondents**

The words of Fig. 10, referring to the fourth question, were written only by male gender and illustrate the thought of men in two age groups 18-29 and 50-69. Inside the word cloud it can be seen the following concepts/words: “troppa sincerità” [“too much sincerity”], “troppa gelosia” [“too much jealousy”], “mancanza di maturità” [“lack of maturity”], “mancanza di condivisione” [“lack of sharing”], “insoddisfazione” [“dissatisfaction”], “distacco” [“apathy”], “falsità” [“deceitfulness”], “avidità” [“greed”], “mancanza di figli” [“childlessness”], “invidia” [“envy”], “inganno” [“deceit”], “la suocera” [“the mother-in-law”] etc.

### G. *Fifth question: "free thought"*

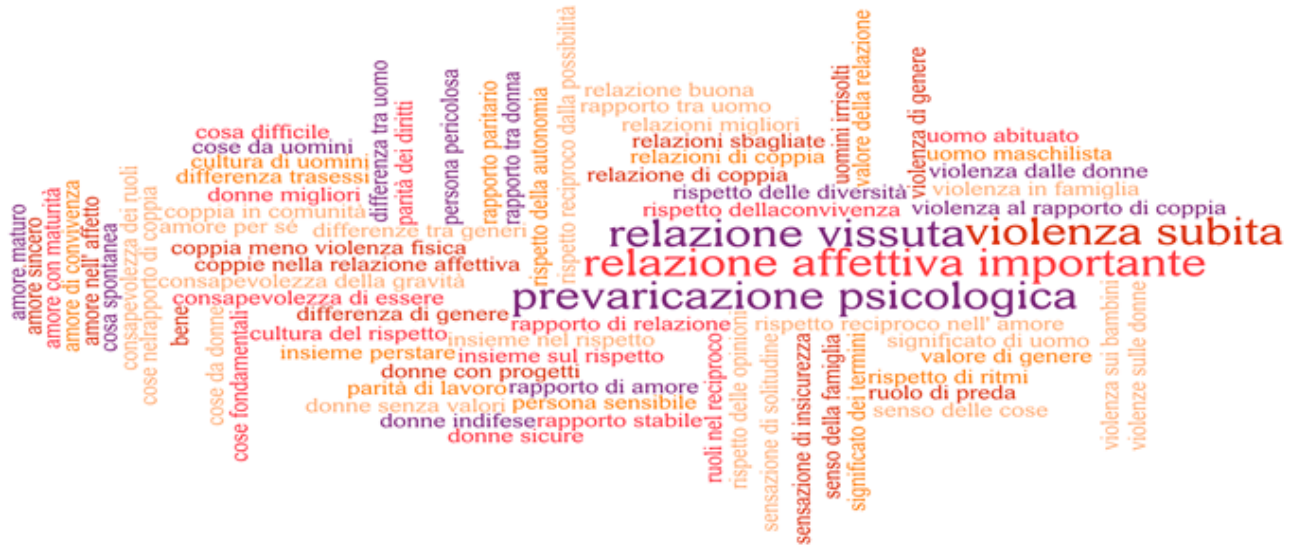

**Figure 11: top rated bag of words for fifth question**

In Fig. 11, it can be seen the answer to the fifth and last question. In this question, the interviewee was given the possibility to write a "free thought": the word cloud contains some groups of words that were among the most frequent.

Not all people, who have completed the questionnaire, responded to the fifth question. The phrases were analyzed and word sequences were extracted. The most frequent words were: “relazione” ["relationship"], “prevaricazione” ["abuse"], “violenza” ["violence"], “amore” ["love"], “rispetto” ["respect"], “rapporto” ["relationship"], “sensazione” ["feeling"]. Each of these words is linked to other forming sequences that give rise to fundamental concepts in the man-woman relationship.

Some examples referring to the above words follow:

- “rapporto” [“relationship”]
  - Experienced relationship
  - important emotional relationship
  - loving relationship
  - wrong relationship
  - good relationship
  - the value of the relationship.
- “prevaricazione” [“abuse”]: psychological abuse.
- “violenza” [“violence”]
  - violence
  - gender violence
  - violence perpetrated by women
  - family violence
  - child abuse
  - violence against women;
- “amore” [“love”]
  - mature love
  - true love
  - love with maturity
  - love of coexistence
  - love in affection
  - love of self
  - love relationship.
- “rispetto” [“respect”]
  - culture of compliance
  - together with respect
  - respecting the autonomy
  - mutual respect in love
  - respect for rhythms of life
  - respect for the opinions
  - respect for coexistence
  - respect for diversity.
- “rapporto” [“relationship”]
  - love relationship
  - stable relationship
  - equal relationship
  - rapport of relationship
  - things in a relationship.
- “sensazione” [“feeling”]
  - feeling of loneliness
  - feeling of insecurity.

## III. CONCLUSION

The language is acquired from early childhood, unconsciously, through social interaction and the environment; it is a communication and information tool. It is unavoidable to the society in which one lives. Language contributes to the definition of the meanings, both individual and collective, of the thoughts and values that we associate with words.

In the "Global Gender Gap Report 2013" compiled by the World Economic Forum, Italy is in 71st place with regard to gender equality.

It is fundamental the use of a language respectful of gender equality to gain an effective overcoming of the woman-man discrimination. Careful reflection on linguistic stereotypes in the media, in textbooks and in daily communication in general, could be able to develop a positive and equal gender identity.

The analysis of language can be a source of reflection on simplifications of the world and on how we perceive it.

This brief analysis on gender differences and relationships, referring to the words used in the questionnaire, supports the hypothesis that language is, surely, the mirror of our thinking, our prejudices and cultural stereotypes and evidence of this work shows that gender stereotypes are still deeply rooted in Italian society.

ACKNOWLEDGMENT

All the authors wish to thank “Comune di Genova” for having made available the data to analyse.

REFERENCES

[1] Rashida Manjoo. “Report of the Special Rapporteur on violence against women, its causes and consequences, Rashida Manjoo”. United Nations General Assembly A/HRC/20/16, 2012.

[2] Kartika Tamara Liotard. “On eliminating gender stereotypes in the EU (2012/2116(INI))”. Committee on Women’s Rights and Gender Equality, 6 December 2012.

[3] “Violenza sulle donne. I giovani come la pensano?”, Regional commission for equal opportunity, Veneto Region, April 2011.

[4] “Rappresentazioni di genere e violenza privata”, report of “Azione e contrasto della violenza sulle donne” project, Province of Parma, January 2009.

[5] “A scuola di pari opportunità”, report of “…così diversi, così uguali…” project, Istituto Scolastico Liceo Scientifico Statale “G.P. Vieusseux”.

[6] “Educare alle relazioni di genere”, Consorzio di solidarietà Con.Sol. Soc. Coop, Chieti.

[7] S. A. Basow, “Stereotypes and roles”, Belmont, CA, US: Thomson Brooks/Cole Publishing Co., 1992.

[8] G. N. Powell, D. A. Butterfield, J. D. Parent, “Gender and Managerial Stereotypes: Have the Times Changed?”, Journal of Management vol. 28 no. 2 177-193, April 2002.

[9] F. Sabatini, “More than a preface”, in “Sexism in Italian Language” by A. Sabatini, Presidency of the Council of Ministers of the Italian Republic, Department of Information and Publishing, Rome, 1993.

[10] Gough, H. G., & Heilbrun, A. B., “The Adjective Check List manual”. Consulting Psychologists Press Inc., Palo Alto, CA, 1983.

[11] I. Briggs Myers, M. H. McCaulley, N. Quenk, and A. Hammer. “MBTI Handbook: A Guide to the development and use of the Myers-Briggs Type Indicator” Consulting Psychologists Press, 3rd edition, Palo Alto, CA, 1998.